\documentclass[web, oneside, onecolumn]{ieeecolor}
\usepackage{generic}
\usepackage{cite}
\usepackage{amsmath,amssymb,amsfonts}
\usepackage{algorithmic}
\usepackage{graphicx}
\usepackage{textcomp}
\usepackage{booktabs}
\usepackage{tabularx}
\usepackage{tabu}
\usepackage{xcolor}
\usepackage{verbatim}
\usepackage{physics}
\usepackage{ragged2e}
\usepackage{hyperref}
\makeatletter
\def\endfigure{\end@float}
\def\endtable{\end@float}
\makeatother

\usepackage[font={sf,md,up}, textfont=small]{subfig, caption}

\usepackage{soul}
\usepackage[table,xcdraw]{xcolor}
\usepackage{diagbox}
\def\BibTeX{{\rm B\kern-.05em{\sc i\kern-.025em b}\kern-.08em
    T\kern-.1667em\lower.7ex\hbox{E}\kern-.125emX}}
\begin{document}
\title{Unobtrusive Pain Monitoring in Older Adults with Dementia using Pairwise and Contrastive Training}
\author{Siavash Rezaei\textsuperscript{1}, Abhishek Moturu\textsuperscript{1,2}, Shun Zhao\textsuperscript{1}, Kenneth M. Prkachin\textsuperscript{4}, Thomas Hadjistavropoulos\textsuperscript{3,5}, and Babak Taati*\textsuperscript{1,2,6}
\thanks{Submitted \today.
This work was supported in part by the Canadian Institutes of Health Research (CIHR), by the AGE-WELL Network of National Centres of Excellence, by the Natural Sciences and Engineering Research Council of Canada (NSERC), and by the Toronto Rehabilitation Institute - University Health Network.}
\thanks{\textsuperscript{1}KITE, Toronto Rehabilitation Institute, University Health Network,\textsuperscript{2}Department of Computer Science, University of Toronto,  \textsuperscript{3}Department of Psychology, University of Regina, \textsuperscript{4}Department of Psychology, University of Northern British Columbia, \textsuperscript{5}Centre on Aging and Health, University of Regina, \textsuperscript{6}Institute of  Biomedical Engineering,
University of Toronto. * Corresponding author: Babak.Taati@uhn.ca}
}
\maketitle

\begin{abstract}
Although pain is frequent in old age, older adults are often undertreated for pain. This is especially the case for long-term care residents with moderate to severe dementia who cannot report their pain because of cognitive impairments that accompany dementia. Nursing staff acknowledge the challenges of effectively recognizing and managing pain in long-term care facilities due to lack of human resources and, sometimes, expertise to use validated pain assessment approaches on a regular basis. 
Vision-based ambient monitoring will allow for frequent automated assessments so care staff could be automatically notified when signs of pain are displayed. However, existing computer vision techniques for pain detection are not validated on faces of older adults or people with dementia, and this population is not represented in existing facial expression datasets of pain. 
We present the first fully automated vision-based technique validated on a dementia cohort. Our contributions are threefold. First, we develop a deep learning-based computer vision system for detecting painful facial expressions on a video dataset that is collected unobtrusively from older adult participants with and without dementia. 
Second, we introduce a pairwise comparative inference method that calibrates to each person and is sensitive to changes in facial expression while using training data more efficiently than sequence models.
Third, we introduce a fast contrastive training method that improves cross-dataset performance. Our pain estimation model outperforms baselines by a wide margin, especially when evaluated on faces of people with dementia. Pre-trained model and demo code available at \url{https://github.com/TaatiTeam/pain_detection_demo} 

\end{abstract}

\begin{IEEEkeywords}
Computer Vision, Dementia, Facial Expression, Older adults, Pain, Deep Learning
\end{IEEEkeywords}
 
\section{Introduction}
\subsection{Motivation}

Pain is common and frequent in old age~\cite{charlton_2005}, but older adults are often underdiagnosed and undertreated for pain~\cite{charlton_2005,Chodosh2004}. This problem is especially serious for people with dementia who are often unable to verbally express or otherwise communicate their experience due to cognitive impairment~\cite{Morrison2000}. Effective and validated assessment approaches, based on observation of non-verbal pain behaviours -- e.g.\ vocalizations, facial and body movements -- are available for this population~\cite{Hadjistavropoulos2014}; but these approaches are not implemented in long-term care (LTC), because significant staff shortages make the ongoing monitoring of expressed pain infeasible~\cite{Hadjistavropoulos2009}. Untreated pain can have serious physical (e.g.\ underlying conditions getting worse) and psychological (e.g.\ agitation and aggression) consequences and addressing pain in a timely manner can have a meaningful positive impact on the quality of life of people residing in LTC. The motivation for this work is to develop an ambient monitoring technology to reliably, automatically, and consistently assess pain in order to improve pain management in LTC and to ultimately provide a better quality of care to older adults.

\subsection{Clinically Valid Assessments of Pain in Dementia}
Two clinically validated metrics for assessing pain in dementia exist: 1) the Prkachin and Solomon Pain Index (PSPI)~\cite{10.1016/j.pain.2008.04.010} and 2) the Pain Assessment Checklist for Seniors with Limited Ability to Communicate-II (PACSLAC-II)~\cite{Chan2014}.


\begin{figure}
  \begin{minipage}[c]{0.3\textwidth}
    \includegraphics[scale=0.08]{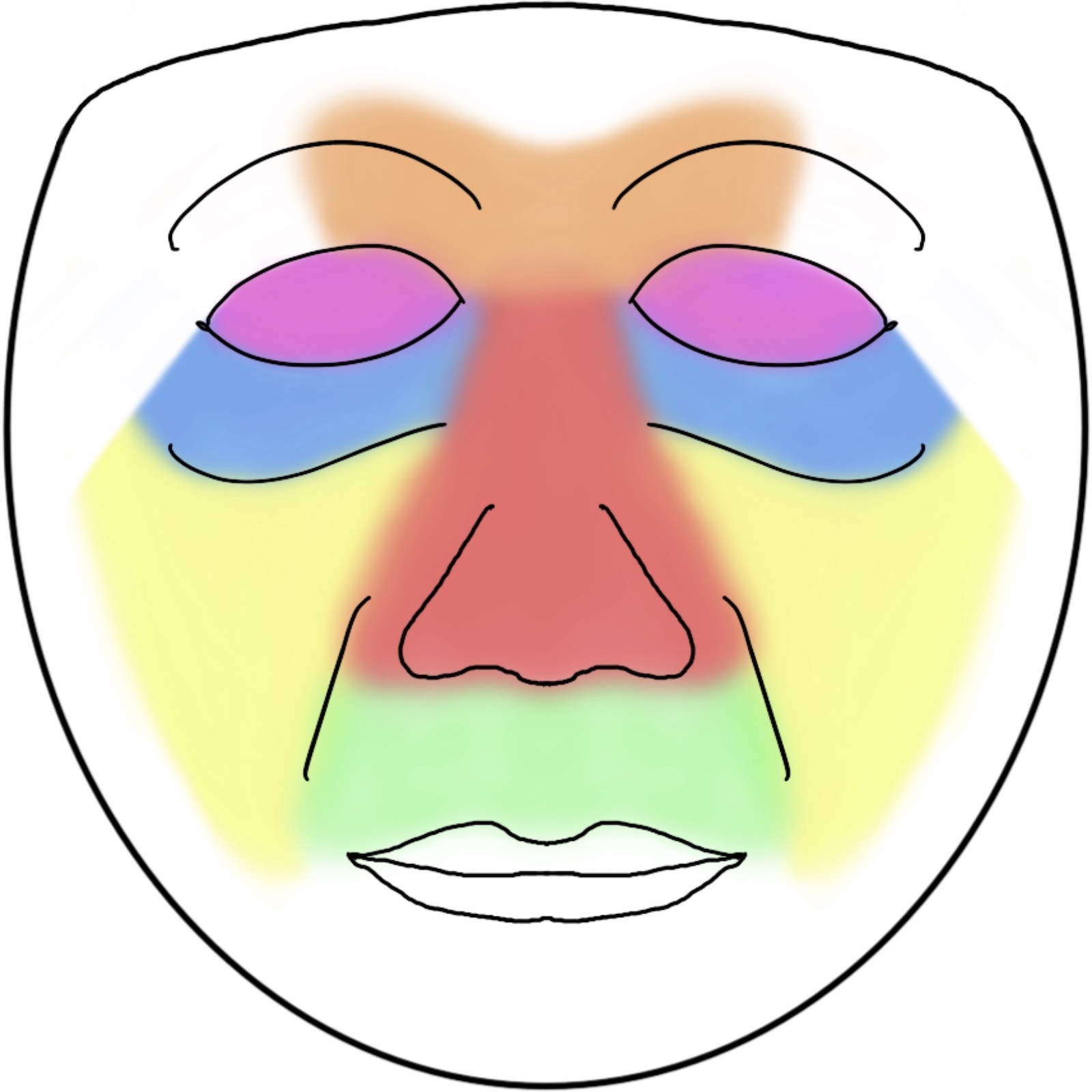}
  \end{minipage}\hfill
  \begin{minipage}[l]{0.7\textwidth}
    \caption{
        The areas of the face where the action units pertaining to PSPI are located: $AU_4$ (orange), $AU_6$ (yellow), $AU_7$ (blue), $AU_9$ (red), $AU_{10}$ (green), and $AU_{43}$ (purple).
    } \label{fig:face_aucs}
  \end{minipage}
\end{figure}

PSPI is a 16-point metric, and is based on the Facial Action Coding System (FACS)~\cite{FACS_manual}. FACS is a detailed taxonomy of different actions units (AU) expressed through facial muscle movements and, as such, provides a good basis for evaluating painful facial expressions. PSPI is computed as follows:
\begin{equation}
PSPI=AU_{43} + max(AU_6, AU_7) + max(AU_9, AU_{10}) + AU_4 \label{eq:PSPI}
\end{equation}
 
\noindent where \(AU_{43}\) corresponds to eyes closed, $AU_{6}$ and $AU_{7}$ correspond to activation of cheek raiser and eyelid tightener muscles, $AU_{9}$ and $AU_{10}$ correspond to levator muscles, and \(AU_{4}\) corresponds to brow lowering muscles (Figure \ref{fig:face_aucs}). Here, $AU_{43}$ is a binary indicator and the others are scored in a $[0,5]$ range.

The PACSLAC-II is a 31-item checklist-based metric designed to have a faster learning curve for annotators. It considers movements of the body and vocalizations in addition to a set of facial expressions. To employ either of these metrics at a care facility, caregivers need to complete the required training. 

Both PSPI and PACSLAC-II are well validated measures for assessing pain in people with dementia~\cite{Kunz2007, 10.1002/ejp.1177}. In this paper, we focus on the automatic estimation of PSPI.

\subsection{Related Work}
\label{subsec:Related Work}
The advances in computer vision in the last two decades have prompted significant volume of work on automatic pain intensity estimation~\cite{8865626}. While other pain datasets exist (e.g.\ the X-ITE database~\cite{gruss2019multi} and the BioVid~Emo~DB~\cite{zhang2016biovid}), the vast majority of this work has been developed and validated using two openly available datasets dedicated to pain estimation, the UNBC-McMaster Shoulder Pain Expression Archive Database~\cite{10.1109/fg.2011.5771462} and the BioVid Heat Pain Database~\cite{Walter2013}. The range of work is diverse in terms of the learning algorithms,
learning tasks, and modalities used.


Ashraf et~al.~\cite{Ashraf2009} used features derived from an active appearance model (AAM) to train a Support Vector Machine (SVM) classifier. They performed pain detection (binary classification) both per-frame and for a sequence of frames. For sequence-level prediction, they aggregated frame level predictions by averaging. On the UNBC-McMaster dataset they achieve an F1 score of 0.56 and 0.48 for per-frame and sequence level prediction, respectively. Interestingly, the performance of their model degrades for sequence level prediction. Also on the UNBC-McMaster dataset, Hammal~and~Cohn~\cite{Hammal2012} applied Log-Normal filter based features and SVMs to classify pain into for four levels of intensity.

Kaltwang~et~al.~\cite{Kaltwang2012} first used each of facial landmarks, Discrete Cosine Transform (DCT), and Local Binary Patterns (LBP) features to train three Relevance Vector Regression (RVR) models separately. They showed that best performance was achieved by training a fourth RVR model to combine the predictions of their three separately trained RVR models. Their method also achieved better performance when they directly predicted PSPI as opposed to predicting the action units and then computing PSPI from them.

More recently, deep learning methods have also been applied to pain estimation directly from images or a sequence of frames~\cite{7849133, 1704.03067, Ma2019, 8486516}. 
Zhou~et.~al~\cite{Zhou2016} used the UNBC-McMaster dataset to train a recurrent convolutional model end-to-end. They used frame sequences of length 30 frames (1 second) with the goal of estimating the pain level in the last frame. They achieved a Pearson correlation of 0.65 with manually annotated pain levels. Egede~et.~al~\cite{Egede2017} fused prediction from pre-trained deep-learned features, Histogram of Oriented Gradients (HOG) features, and facial landmarks using RVR models. They achieved a Pearson correlation of 0.67 on the UNBC-McMaster dataset. Tavakolian~\&~Hadid~\cite{8545324, Tavakolian2019} used 3D convolutions to model temporal aspects of videos in their two recent works. In \cite{8545324} and \cite{Tavakolian2019}, they report a Pearson Correlation Coefficient (PCC) of 0.83 and 0.92 on the UNBC-McMaster dataset, respectively.

The UNBC-McMaster dataset is a unimodal dataset and is used for developing and evaluating vision-based pain assessment techniques. The BioVid, BioVid~Emo~DB, and X-ITE datasets, by contrast, are multimodal and their availability has sparked extensive interest in multimodal pain detection~\cite{kachele2015multimodal,werner2019twofold}.  

It is also worth noting the rich body of work focused on pain detection in neonates. In one of the earliest works, Brahnam~et~al.~\cite{Brahnam2006} developed the first pain classification dataset, the Infant Classification of Pain Expressions (COPE) database. They used Principal Component Analysis (PCA), Linear Discriminant Analysis (LDA), and SVM to discriminate between induced pain and other stressors.
More recently, Celona~\&~Manoni~\cite{Celona2017} used a mixture of hand crafted features such as LBP and HOG and features extracted from a pre-trained convolutional neural network. They then applied PCA and trained a linear SVM to perform pain assessment in the COPE database. More recently, Salekin~et~al.~\cite{8914537} proposed a fully deep learning-based model which takes into account both facial expressions of neonates as well as their body movements. They also model the temporal aspect of the task using a recurrent neural network. They evaluate their algorithm on video data collected from neonates at Neonatal Intensive Care Units (NICU) using the NIPS~\cite{HudsonBarr2002} pain scale.





\subsection{Contributions}

Our work has three main contributions, addressing gaps in existing work, as detailed below. In experimental analysis, we show that our method outperforms baselines by a wide margin specially on participants with dementia. 

\subsubsection*{Fully Automated Pain Detection in Dementia}
Although there has been extensive work on vision-based pain detection~\cite{Hassan2019}, work focusing on detecting pain in older adults with dementia, who stand to benefit the most from this technology, is scarce~\cite{1811.07988}. Atee~et~al.~\cite{AteeHotiParsonsHughes2017}  developed a system for pain assessment specifically for people with dementia, but their system requires manual entry of observations with an operator through a mobile app. Previous work has shown facial analysis models that work well on faces of healthy young adults do not necessarily yield similar performance when evaluated on faces of older adults with dementia~\cite{8643365}. Our first contribution is the development and evaluation of a computer vision model to fully automatically and unobtrusively estimate pain from facial expressions of older adults with dementia. To our knowledge, our work is the first of its kind to employ these techniques on a relatively large dataset that is collected unobtrusively from the target population.

\subsubsection*{Pairwise Pain Detection}
Our model was developed by drawing on insights from how trained human annotators perform the same task. It has long been recognized that incorporating temporal information leads to significant gains in performance in automatic facial expression recognition. As a result, many models have been developed to take advantage of temporal information in a sequence of frames in videos~\cite{7849133, 1704.03067, Ma2019, 8486516}. However, given the small size of the datasets available for facial expression recognition, the number of distinct sequences (i.e. training samples) is very small compared to the situation where each frame is treated as a training sample. Our second contribution is the use of pairwise pain detection to address this issue. We propose an architecture that takes two frames as input: a reference frame and a target frame. With this method, the number of samples is not only uncompromised, but expanded, because there are more possible pairings than there are individual frames. At the same time, our method captures what is good about using sequences, namely analysing \emph{changes} in expression from frame to frame.

\subsubsection*{Contrastive Training}
Contrastive training which was originally developed to train probabilistic and energy-based models \cite{Hinton1999, Hinton2002}, has had a resurgence in deep learning and been shown to improve classification metrics \cite{2002.05709, 2004.11362, Grathwohl2019}. Our third contribution is the introduction of a contrastive training method in a regression setting, and a fast method for generating out-of-distribution samples for image data. We show that this approach improves within-dataset and cross-dataset performance.

\subsection{Organization}
The remainder of this paper is organized as follows. Data used to develop and validate this model is reviewed in Section~\ref{sec:Data}. The details of the method and comparison baselines are presented in Section~\ref{Method}. The results of evaluations on both cognitively healthy control and participants with dementia are reported and discussed in Section~\ref{Results}. Ethical considerations and further work required to deploy this model in a clinical setting are presented in Section~\ref{ethic} and Section~\ref{conclusion}, respectively.

\section{Data}
\label{sec:Data}
\subsection{Datasets Used}
\label{sec:Datasets}
We used two datasets to conduct our experiments, the University of Regina (UofR) Pain in Severe Dementia dataset~\cite{10.1002/ejp.1177} and the UNBC-McMaster Shoulder Pain Expression Archive Database~\cite{10.1109/fg.2011.5771462}. The publicly available portion of the UNBC-McMaster dataset contains video data from 25 participants (13 females) with a shoulder injury during painful and non-painful movements, recorded at 30 frames per second (fps) and in Quarter VGA (240$\times$320) resolution. The videos are manually annotated with FACS codes so that the PSPI score can be calculated for each video frame. The dataset contains 48,391 image frames in total (1936~±~837 per participant). 

The UofR dataset contains video data from 102 older adult participants with and without dementia, recorded at 15 fps. In our experiments, UofR videos were transcoded to VGA (640$\times$480) resolution. Each session was filmed with three synchronized cameras recording top, left, and right views of the face. The videos were first recorded during a baseline state when the participant was lying on a bed and then during an examination state in which a licensed physiotherapist assisted the participant to execute a sequence of movements to identify painful areas. Videos of 95 people from the dataset (74 females) were annotated manually by trained annotators according to PSPI and PACSLAC-II pain rating scales. Of these 95 older adults, 47 were community dwelling and cognitively healthy (age: 75.5~±~6.1), whereas the remaining 48 (age: 82.5~±~9.2) were individuals with severe dementia residing in LTC. 

\begin{figure}[!t]
\centerline{\includegraphics[width=\columnwidth]{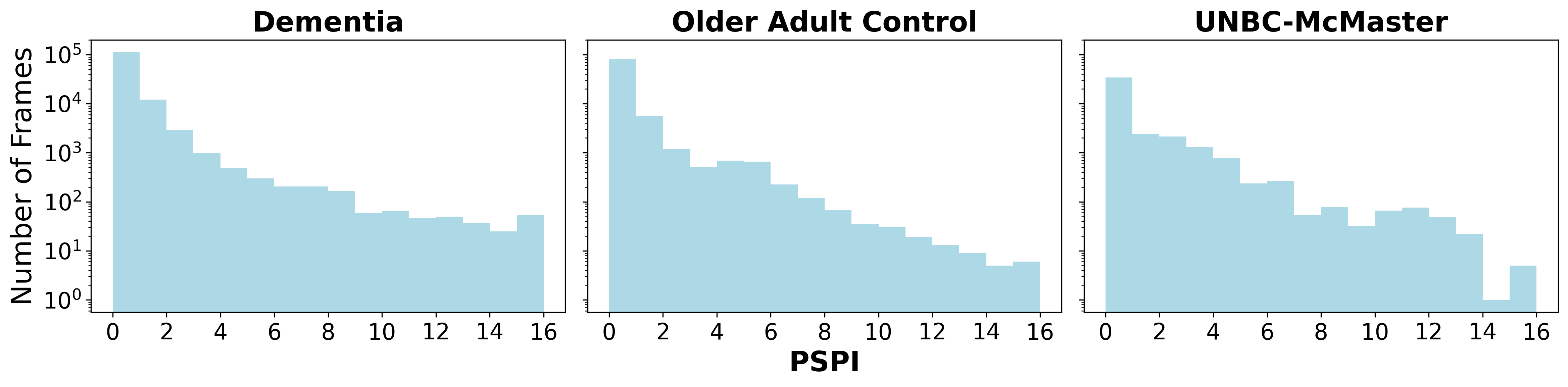}}
\caption{Number of frames at each pain level (PSPI score) in the UofR dataset (separated by cognitive status) and in the UNBC-McMaster dataset. Note that the y-axis is in log scale.}
\label{pspi_dist}
\end{figure}

Figure~\ref{pspi_dist} depicts the distribution of pain levels (PSPI) in each dataset. As the figure illustrates, the distribution is highly skewed, and frames with zero or low PSPI occur significantly more frequently than those with a high PSPI. For instance, the ratio of frames with PSPI\,$>$\,5 to those with PSPI\,$<$\,2 is 1.8\% for the UNBC-McMaster dataset, and 0.8\% and 0.7\% for the dementia and healthy control portions of the UofR dataset, respectively. 

Both datasets used in this paper involved human participants and were collected in accordance with the ethical standards and approval of the respective institutional review boards and with the Helsinki declaration. All participants, or substitute decision makers in cases of severe dementia, provided informed consent. The UofR protocol was approved by the institutional review board at the University of Regina (REB\# 2014-132) and at the University Health Network (REB\# 15-9342-DE)

\subsection{Data Preprocessing}
The same preprocessing steps were applied to the video data from both the UofR and the UNBC datasets. The main aim with preprocessing was to remove variability in the data that was known not to be related to facial actions, thereby limiting the opportunity for models to overfit. As a first step, frames were cropped around the face, using the single shot scale-invariant face detection (S3FD) model~\cite{S3FD}. Frames in which S3FD did not detect the face (e.g.\ because of partial occlusion, poor lighting, extreme angles) were not further processed. 

A challenge in preprocessing was the high variability in head pose and camera angle across different participants and over the length of each video. This was particularly the case for the UofR dataset, where the recording protocol involved movements such as transitioning from lying down to sitting up. In order to remove this variability, 68 facial landmarks were extracted using the Face Alignment model (FAN)~\cite{bulat2017far}. The landmarks were then used to create a frontal and upright rendering of the face. This approach also normalized variance due to differences in face geometry. Two methods were experimented to create the frontalized images. The first method was a piecewise affine transform and the second method involved fitting a face-shaped 3D mesh~\cite{Huber2016} onto the face and then projecting the image onto this mesh. The second method avoided some of the distortions introduced by the piecewise affine transformation, especially at extreme angles. However, while being computationally more expensive, it did not result in performance improvements. As such, all experimental results reported in this paper are based on using the piecewise affine transform. 

It was observed that the accuracy of the FAN landmarks rapidly declined as the participants' head pose deviated from facing the camera. The misplacement of landmarks resulted in heavily distorted frontal renderings. To address this, a small subset of the frames (1740) were randomly selected and manually annotated with a binary front view vs. non-front view label. This dataset was subsequently used to train a model to assign a frontal score to each frame. This model was then used to discard profile view faces from the UofR dataset.
After discarding the frames in which S3FD did not detect the face and the frames which had a low frontal score, the total number of frames used from the UofR dataset was 162,629 (65\% of total).

Videos were recorded at vastly different lighting conditions and camera settings. In order to remove this source of variability, Contrast Limited Adaptive Histogram Equalization~\cite{zuiderveld_1994} was applied to each frame.
\section{Method}
\label{Method}

\subsection{The Proposed Model}

\begin{figure*}[!t]
\centerline{\includegraphics[scale=0.621]{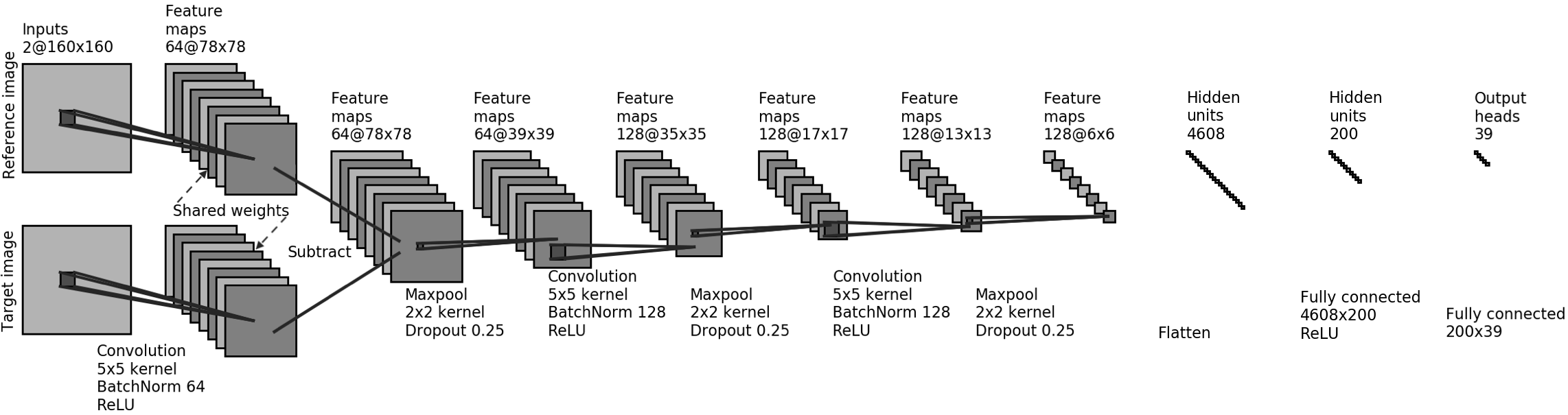}}
\caption{The proposed model architecture.}
\label{model_arch}
\end{figure*}

\subsubsection*{Architecture}

Our primary aim was to train a model to predict pain level at each image frame such that estimated PSPI scores correlated strongly with trained human annotations. The model proposed is based on the observation that human annotators: 1) adjust to each person's resting face, and 2) are sensitive to movements~\cite{Bassili1979}. To integrate this insight into the model without adding too much complexity, a convolutional model was used to take two images from each participant as input: a reference image and a target image. The goal of the training process is for the network to estimate the difference in FACS codes and PSPI intensities between the target and the reference image. To get the predicted output for the target frame, we add the model’s predicted difference to the reference frame’s ground truth.

Figure \ref{model_arch} depicts the network architecture. The first layer of the network is applied to both reference and target frames independently and the resulting feature maps from each frame are subtracted. The rest of the network follows the typical motif of convolutional network architectures (resembling LeNet~\cite{726791}) where a convolution operation is followed by an affine batch-normalization layer and ReLU activation. The output of each convolutional layer goes through a max-pooling layer and a dropout layer. The last convolution layer is followed by two fully connected layers.

\subsubsection*{Multi-task Learning}

The model was trained in a multi-task manner. That is, all action units and PSPI scores from each dataset were included as separate target variables for the model to estimate. Table~\ref{tab:output_heads} indicates which annotations were available and used for each dataset. For the UofR dataset PACSLAC-II annotations were also available, and therefore, PACSLAC-II action units corresponding to facial actions were also included as targets. As the model was trained on pairs of images (reference and target images), multi-task learning meant estimating the difference between FACS AUs, PSPI, and PACSLAC-II annotations.  
Each training sample was created by randomly selecting two frames from the same person. At each epoch a fresh set of pairs was created as a data augmentation strategy. When the model is deployed in a clinical setting, only a rest state reference image of the person can be obtained, e.g.\ an image capture from each LTC resident at the time of admission. Since the resting image will be in a no pain (PSPI=0) condition, the delta PSPI will always be non-negative. In light of this, the loss function was designed to generate a learning signal only when the target delta PSPI was non-negative. This means for all outputs the target is always bigger than reference during training. At test time, only frames with a PSPI score of zero are chosen as reference frames.

\begin{table}[!thp]
\begin{tabularx}{\columnwidth}{l|>{\RaggedRight}X>{\RaggedRight}X>{\RaggedRight}X}
\toprule
 \diagbox{Annotations}{Datasets\\ \\ \\}& Dementia & Older Adult Control & UNBC-McMaster\\
\midrule
PSPI & \checkmark  &  \checkmark &  \checkmark \\
FACS AU43 &\checkmark  &  \checkmark &  \checkmark \\
FACS AU4  &\checkmark  &  \checkmark &  \checkmark \\
FACS max(AU9, AU10) &\checkmark  &  \checkmark &  \\
FACS max(AU6, AU7)  &\checkmark  &  \checkmark &  \\
PACSLAC-II P1  &\checkmark  &  \checkmark &  \\
PACSLAC-II P2 &\checkmark  &  \checkmark &  \\
PACSLAC-II P3 &\checkmark  &  \checkmark &  \\
PACSLAC-II P4 &\checkmark  &  \checkmark &  \\
PACSLAC-II P5 &\checkmark  &  \checkmark &  \\
PACSLAC-II P6 &\checkmark  &  \checkmark &  \\
PACSLAC-II P7 &\checkmark  &  \checkmark &  \\
PACSLAC-II P8 &\checkmark  &  \checkmark &  \\
PACSLAC-II P9 &\checkmark  &  \checkmark &  \\
PACSLAC-II P10 &\checkmark  &  \checkmark &  \\
PACSLAC-II P11 &\checkmark  &  \checkmark &  \\
FACS AU9  &  &   &  \checkmark \\
FACS AU10  &  &   &  \checkmark \\
FACS AU6  &  &  &  \checkmark \\
FACS AU7  &  &  &  \checkmark \\
\bottomrule
\end{tabularx}
\caption{Check marks indicate that the annotation was available for that dataset and was used as a target during training. There are 39 check marks corresponding to the 39 outputs of the network}
\label{tab:output_heads}
\end{table}
\subsubsection*{Contrastive Training}
Grathwohl~et~al.~\cite{Grathwohl2019} note that there is an additional (unused) degree of freedom in the softmax function of neural network based classifiers. They train that degree of freedom contrastively to distinguish between in-distribution (ID) and out-of-distribution (OOD) samples. Their experiments showed that when a classifier is trained in this manner, it develops desirable properties such as better calibration, robustness to adversarial perturbations, and the ability to detect OOD samples.
Inspired by their findings, here, a contrastive training method is developed for regression. We note that the output of a regression neural network model is the dot product between the flattened output (features) of the penultimate layer and the weights of the last fully connected linear layer. The dot product depends on both the magnitudes and cosine of the angle between the two vectors. Here, an additional contrastive loss term is added as follows, constraining the direction (but not the magnitude) of the feature vectors:

\begin{equation}
Loss_{contrastive}=
\frac{1}{col}\sum_{i}^{col}\abs{\frac{f(\hat{x}) \cdot W_{fc}[i]}{\norm{f(\hat{x})}\norm{W_{fc}[i]} }} - \abs{\frac{f(x) \cdot W_{fc}[i]}{\norm{f(x)}\norm{W_{fc}[i]} }} \\
\label{cont_loss}
\end{equation}

\noindent where \(\hat{x}\) represents an OOD sample, \(x\) represents an ID sample, \(f\) represent the network up to (but excluding) the last linear layer,  \(W_{fc}\) represents the weights of the last fully connected linear layer,   {and \textit{col} is the number of columns in $W_{fc}$}. The sum is over the columns of \(W_{fc}\). Minimizing this loss term pushes the cosine of the angle between the feature vectors and the last layer's weights towards zero for OOD samples, and pushes it away from zero for ID samples. Total loss is then the regression loss -- mean squared error in this case -- plus the scaled contrastive loss:
\begin{equation}
Loss_{total}= Loss_{regression} + c \cdot Loss_{contrastive}
\label{total_loss}
\end{equation}

In \cite{Grathwohl2019}, OOD samples are generated using Stochastic Gradient Langevin Dynamics (SGLD)~\cite{Welling-Teh2011}. This method of generation requires an inner loop within the training loop for the OOD samples to converge. This slows down training significantly. To avoid this computational cost, we used random distortions instead to generate OOD samples. Images were distorted aggressively such that they no longer resembled an upright face, but did preserve characteristics of natural images such as local structures. The distortions that were used here were flipping 
and elastic transformation~\cite{10.1109/icdar.2003.1227801}. This method of OOD sample generation does not fully follow the logic of contrastive divergence employed in energy-based model training. However, we intuit that by starting from ID images and distorting them, we obtain OOD samples that are similar enough to ID samples to provide a useful learning signal.
The results are reported with and without contrastive training.

\subsubsection*{Augmentation and Hyper-parameters}
Experiments were run using several data augmentation techniques. Best results were obtained by combining random crops and horizontal flipping. The model was trained with a constant learning rate and weight decay of $1e-4$ with the Adam optimizer for 70 epochs. Batch sizes were 32 and dropout probability was 0.25 for all dropout layers. The coefficient \(c\) for contrastive training was set to 0.05.

\subsection{Baseline Models}

To establish a baseline, we used three methods. The first baseline was the method proposed by Kaltwang~et~al.~\cite{Kaltwang2012}, in which a Relevance Vector Regression (RVR) model is trained on the discrete cosine transform (DCT) of each frame to estimate the PSPI. Here Support Vector Regression (SVR) was used instead of RVR.

The second baseline was the openly available OpenFace library; a facial behavior analysis toolkit which provides FACS action unit detection functionality from videos. OpenFace takes a non-deep learning approach proposed by Baltrusaitis~et~al.~\cite{Baltrusaitis2015}. It uses HOG features and facial landmarks as predictors and trains an SVR model to infer FACS codes. To adapt to individual differences across people, they calibrate to the face of each person by computing a neutral face as the mean of the features across a sequence of frames in the video. The predicted FACS codes from OpenFace were then used to compute the PSPI pain score. OpenFace does not predict $AU_{43}$ (eyes closed), but it does predict $AU_{45}$ (blink). We used $AU_{45}$ instead of $AU{43}$; so there is a small deviation from the true definition of PSPI.

The third baseline was the model proposed by Pau~et~al.~\cite{7849133}. Here, we replicated their method by first fine-tuning a pre-trained VGG-Face~\cite{Parkhi15} model with our data, and then training an LSTM with features extracted from FC6 layer of the said VGG model. The reasoning behind this approach is that the LSTM is able to model temporal relationships between video frames, and thereby improve performance over models that estimate pain from a single frame. It is worth pointing out that due to considerable size and computational requirements of VGG16, using this model for real-time analysis of video data is not currently feasible.

The fourth baseline was a convolutional neural network (CNN) model developed by Ertugrul et~al.~\cite{Ertugrul2019} as part of an automated facial affect recognition (AFAR) tool. The AFAR model was shown to be effective in predicting FACS probabilities. Here the AFAR convolutional model was trained with a minor modification to convert it to a regression model rather than a classification model. The FACS predictions were then used to compute PSPI.
We also evaluated a commercially available software (FaceReader, by Noldus Information Technology, Netherlands), but the results are not reported due to poor performance and thus not constituting an appropriate baseline.

\subsection{Cross Validation}
To perform validation and compare performance, the data was divided into 5 folds with some additional constraints. First, the folds were constructed such that all samples (i.e.\ video frames) from each participant were either in the training or in the validation set; i.e.\ leave-k-subjects-out cross validation. Second, the proportion of the dementia and healthy control people were kept the same between training and validation sets. The number of samples across folds varied as different people had different numbers of samples. The same folds were used across all experiments to ensure that the results were comparable. 
\begin{table*}[!thp]
\centering
\begin{tabular}{l|ll|ll|ll|ll}
\toprule
\textbf{Fold} &\multicolumn{2}{l}{\textbf{Dementia}} & \multicolumn{2}{l}{\textbf{Older Adult Control}} & \multicolumn{2}{l}{\textbf{UNBC-McMaster}} & \multicolumn{2}{l}{\textbf{Total}}\\
&frames(K)&subjects&frames(K)&subjects&frames(K)&subjects&frames(K)&subjects \\
\midrule
1 &27 / 116&10 / 38&20 / 77&10 / 37& 11 / 37&5 / 20&59 / 230&25 / 95 \\
2 &34 / 109&10 / 38&13 / 84&8 / 39& ~8 / 40&5 / 20&56 / 233&23 / 97 \\
3 &29 / 114&10 / 38&23 / 74&10 / 37& ~9 / 39&5 / 20&62 / 227&25 / 95 \\
4 &22 / 121&~8 / 40&19 / 78&9 / 38& 11 / 37&5 / 20&53 / 236&22 / 98 \\
5 &30 / 113&10 / 38&21 / 76&10 / 37& ~8 / 40&5 / 20&58 / 231&25 / 95\\
\bottomrule
\end{tabular}
\caption{Number of test / train frames and subjects in each fold broken down by dataset. Note: number of frames is rounded to thousands.}
\label{tab:fold_statistics}
\end{table*}

\subsection{Binary Classification}

An important use case of this model in a clinical environment is to notify the caregiving staff when painful expressions are detected. The decision to notify the staff can be framed as a classification problem. To derive class predictions, the output of the best performing regression model is thresholded. The decision threshold for pain was selected to be 2 and above. The rationale for this threshold is discussed in Appendix~I.

\subsection{Evaluation Metrics}

To evaluate the quality of predictions on the test data, two types of metrics were used. In regression, the performance was quantified via the Pearson correlation between the predictions and the human annotated ground-truth. In binary classification, to distinguish pain vs. no-pain frames, F1 score, average precision, and Area Under the ROC Curve (AUC) were used to quantify the performance.

\section{Results and Discussion}
\label{Results}
Table~\ref{tab:3_results} reports the results of six experiments. The first row reports the per-frame Pearson correlation coefficients between our model's predictions and ground truth annotations. 
One benefit of the proposed model is that multiple predictions can be made for a target frame by using different reference frames. Those predictions can then be averaged to obtain a more accurate prediction. Note that in all the results reported throughout the paper, the reference frame is always selected to have a PSPI score of 0. The second row in Table~\ref{tab:3_results} reports the results when the prediction for each target frame is obtained by averaging 5 predictions from pairing that target frame with 5 different reference frames (of the same person).
The third row of Table~\ref{tab:3_results} reports the results for the case where the reference and target frame pairs were not constrained to be from the same person \emph{during training}. 
Comparing the second and third rows of the table, we see that the model performs significantly better when it is \emph{trained} with reference and target frame pairs from the same person, as opposed to when the reference frames are from randomly picked people. Note that both rows are reporting \emph{test} results where the reference and target frames \emph{are} from the same person.
This provides evidence that by making sure that the reference and target frames are from the same person during training, the model learns to use the reference frame to calibrate itself to each individual. Note that all other results reported in the paper are from models that were trained with reference frames from the same person. Moreover, all reported test results are calculated by averaging the predictions obtained using five different reference frames.
Finally, Table~\ref{tab:3_results} compares the performance of our model with vs. without contrastive training. 

\newcolumntype{M}[1]{>{\centering\arraybackslash}m{#1}}
\newcolumntype{N}{@{}m{0pt}@{}}
\begin{table*}[thp]
\centering
\begin{tabular}{m{8.1cm}M{2.32cm}M{2.32cm}M{2.32cm}N}
\toprule
Experiment & Dementia & Older Adult Control & UNBC-McMaster\\
\midrule
Trained with reference frames from the same person &0.38 / 0.45 & 0.54 / 0.56 & 0.69 / 0.68 \\[2pt]
Trained with reference frames from the same person (mean of 5) &0.42 / 0.48 & 0.58 / 0.58 & 0.71 / 0.69 \\[2pt]
Trained with reference frames from random persons (mean of 5) &0.29 / 0.32 & 0.35 / 0.36 & 0.58 / 0.61 \\
\bottomrule
\end{tabular}
\caption{Pearson correlations for per-frame predictions of different experiments; without\,/\,with contrastive training. Note: In all of the rows, the test results are obtained by using reference frames from the same person. For rows 2 and 3, the mean of 5 predictions was used, where the 5 predictions were obtained by using 5 different reference frames.
}
\label{tab:3_results}
\end{table*}

The gap in performance of the models between the UofR and UNBC-McMaster datasets, highlights the difficulty of the task in unobtrusive and less controlled environments; as the UofR dataset was collected in less controlled conditions, both in terms of lighting and camera positioning. This is in contrast to UNBC-McMaster dataset where people face the camera and are well lit. This gap in performance could also, at least partially, be attributed to the difficulty in distinguishing expressions of pain in older faces in the UofR dataset. The large gap in performance between the dementia and healthy control cohorts of the UofR dataset could be primarily attributed to the difficulty in distinguishing expressions of pain in faces of people with dementia. 

Our proposed model has three PSPI outputs, one for each of UofR dementia, UofR healthy control, and UNBC-McMaster. Each output is only trained on samples from its corresponding dataset. To support the assertion about the difficulty of the UofR dataset, cross-dataset performance was measured. 
That is, each of the three PSPI outputs were used to get predictions for samples from all three datasets. This is similar to the analysis performed by Othman~et~al.~\cite{othman2019cross}, who performed cross-database evaluation of vision-based pain recognition models trained on the BioVid and X-ITE datasets. The difference is that, in our case, a single model is concurrently trained on all three datasets, but each output is only trained with samples from the corresponding dataset. For instance, if a training sample is from the UNBC dataset, only outputs corresponding to UNBC contribute to the loss. These results can be seen in Table~\ref{tab:cross_dataset}. 
One can see that both the dementia and control outputs perform significantly better on the UNBC-McMaster dataset than their own datasets, while the UNBC-McMaster output performs worse on both the dementia and control datasets.
Furthermore, it can be seen that contrastive training not only generally boosts performance (Table~\ref{tab:3_results}), but also helps in reducing the performance variation on each dataset across outputs (Table~\ref{tab:cross_dataset}). This suggests that contrastive training may help the model to generalize better to unseen data.

\begin{table}[!thp]
\begin{tabularx}{\columnwidth}{l|>{\RaggedRight}X>{\RaggedRight}X>{\RaggedRight}X}
\toprule
 \diagbox{Dataset}{Output\\head}& Dementia & Older Adult Control & UNBC-McMaster\\
\midrule
Dementia &\textcolor{gray}{0.42 / 0.48} & 0.43 / 0.49 & 0.39 / 0.48 \\
Older adult control &0.48 / 0.54 & \textcolor{gray}{0.58 / 0.58} & 0.51 / 0.57 \\
UNBC-McMaster&0.68 / 0.69& 0.69 / 0.70 & \textcolor{gray}{0.71 / 0.69}\\
\bottomrule
\end{tabularx}
\caption{Cross-dataset Pearson correlations of per-frame PSPI predictions; without/with contrastive training. }
\label{tab:cross_dataset}
\end{table}

Tables~\ref{tab:pearson_results}~and~\ref{tab:f1_results} compare regression and classification results from our proposed model to those of the baseline models. The results are reported for both per-frame predictions and rolling windows. The rolling window results are reported for three different lengths of rolling windows. The rolling windows have more ecological validity than individual frames in that in a clinical setting judgements about pain are made by observing the person for a period of time. For each window, predictions were aggregated by taking the maximum. As shown, in both regression and classification tasks, further improvement can be achieved on the UofR dataset by this aggregation.


In order to make the comparison to the AFAR model more fair, another experiment was performed where the AFAR model was trained and tested on images where the mean face of each person was subtracted from every image of that person. This is a way of amplifying deviations from a persons resting face. The results of this experiment are reported on the fourth rows of Tables~\ref{tab:pearson_results}~and~\ref{tab:f1_results}. We can see an improvement in performance, specifically for the healthy control participants. However, the performance still lags behind our proposed model, especially on participants with dementia.

The VGG16+LSTM model proposed in~\cite{7849133} is similar to AFAR (mean-subtracted) and also our model in that it uses more information than a single frame to estimate pain. Particularly, this model uses the previous 9 frames plus the current frame in order to estimate pain for the current frame. However, it performs rather poorly, except for the classification task on the UNBC dataset, as can be seen in Tables~\ref{tab:pearson_results}~and~\ref{tab:f1_results}. We attribute this poor performance to over-parameterization which resulted in severe overfitting. We alleviate the overfitting problem by using regularization and early-stopping, however, this did not fully eliminate the problem.
It is worth noting that this model has $30\times$ more parameters than our proposed model. Moreover, its computational requirements make it infeasible for real-time pain detection. We were only able to obtain a PCC of 0.48 when we replicated \cite{7849133}, which is significantly lower than their reported PCC of 0.78. We believe that this discrepancy is due to the fact that the validation set in this work is balanced following recommendation from \cite{6681438}. However, here we are computing PCC from unbalanced test data. We believe balancing the validation set defeats the purpose of cross-validation, because the purpose of cross-validation for us is to estimate the performance in a test scenario, i.e. actual deployment of the algorithm. By balancing the validation set, we would change the data distribution even further away from what is expected in a test scenario. Therefore, performance statistics on a balanced validation set do not provide a good estimate of test performance in real deployment.

Comparing the two last rows (our models) with the rest (baselines) in Tables~\ref{tab:pearson_results}~and~\ref{tab:f1_results} shows that pairwise prediction is superior to single-frame prediction methods. We emphasize that this gain cannot solely have originated from the models themselves, but rather due to pairwise training and inference. We base this inference on the fact that AFAR and our proposed model have about the same number of parameters and a very similar architecture.

For the Regina dataset, videos  from a random subset of 25 participants were selected and annotated by a second rater in an identical procedure to the first raters. This inter-rater agreement (quantified by the PCC between the first and second raters) is reported in the last row to serve as an upper bound for achievable performance. The UNBC-McMaster dataset similarly reports the Pearson correlation between the first and second raters. However, this correlation was not computed for PSPI, which is the pain scale of interest in this work.
\begin{table*}[!thp]
\centering
\begin{tabu}{lllll|llll|llll}
\toprule
\textbf{Model} &\multicolumn{4}{l}{\textbf{Dementia}} & \multicolumn{4}{l}{\textbf{Older Adult Control}} & \multicolumn{4}{l}{\textbf{UNBC-McMaster}}\\
&frame&1sec&5sec&20sec&frame&1sec&5sec&20sec&frame&1sec&5sec&20sec \\
\midrule
SVR with DCT features &0.17&0.22&0.28&0.39& 0.18&0.22&0.29&0.36 & 0.50&0.55&0.37&N/A \\
OpenFace&0.18&0.21&0.27&0.42 & 0.18&0.20&0.20&0.12 & N/A&N/A&N/A&N/A \\
VGG16+LSTM&0.23&0.24&0.29&0.27 & 0.20&0.20&0.22&0.22 &0.48&0.43&0.27&N/A \\
AFAR &0.31&0.40&0.51&0.60 & 0.32&0.45&0.56&0.59 & 0.58&0.68&0.57&N/A \\
AFAR (mean subtracted)&0.33&0.42&0.53&0.58 & 0.50&0.58&0.62&0.66 & 0.67&0.71&0.57&N/A\\
Pairwise (ours) &0.42&0.47&0.58&0.79 & \textbf{0.58}&0.59&0.63&0.68 & \textbf{0.71}&\textbf{0.76}&\textbf{0.67}&N/A \\
Pairwise w/ contrastive training (ours) &\textbf{0.48}&\textbf{0.57}&\textbf{0.69}&\textbf{0.82} & \textbf{0.58}&\textbf{0.62}&\textbf{0.69}&\textbf{0.70} & 0.69&0.75&0.63&N/A \\
\midrule
Second rater &0.91&0.93&0.95&0.97 & 0.91&0.92&0.94&0.95 &\multicolumn{4}{c}{Data not available} \\
\bottomrule
\end{tabu}
\caption{Regression Pearson correlations of the models and the second rater annotations.}
\label{tab:pearson_results}
\end{table*}

\begin{table*}[!thp]
\centering
\begin{tabu}{lllll|llll|llll}
\toprule
\textbf{Model} &\multicolumn{4}{l}{\textbf{Dementia}} & \multicolumn{4}{l}{\textbf{Older Adult Control}} & \multicolumn{4}{l}{\textbf{UNBC-McMaster}}\\
&frame&1sec&5sec&20sec&frame&1sec&5sec&20sec&frame&1sec&5sec&20sec \\
\midrule
SVR with DCT features &0.14&0.22&0.32&0.47& 0.15&0.23&0.39&0.62 & 0.48&0.57&0.79&N/A \\
OpenFace&0.18&0.21&0.36&\textbf{0.64} & 0.23&0.26&0.34&0.59 & N/A&N/A&N/A&N/A \\
VGG16+LSTM&0.19&0.24&0.23&0.43 & 0.23&0.27&0.26&0.60 &\textbf{0.59}&\textbf{0.65}&\textbf{0.85}&N/A \\
AFAR &0.20&0.28&\textbf{0.45}&0.58 & 0.25&0.34&0.48&0.67 & 0.57&0.62&0.80&N/A \\
AFAR (mean subtracted)&0.23&0.27&0.35&0.52 & 0.43&0.45&0.56&0.70 & \textbf{0.59}&0.63&\textbf{0.85}&N/A\\
Pairwise (ours) &\textbf{0.30}&0.35&0.36&\textbf{0.64} & 0.47&0.47&0.54&\textbf{0.75} & 0.56&0.62&0.79&N/A \\
Pairwise w/ contrastive training (ours) &\textbf{0.30}&\textbf{0.38}&0.42&0.63 & \textbf{0.48}&\textbf{0.49}&\textbf{0.61}&\textbf{0.75} & 0.54&0.63&0.82&N/A \\
\midrule
Second rater &0.74&0.77&0.82&0.87 & 0.90&0.90&0.90&0.90 & \multicolumn{4}{c}{Data not available} \\
\bottomrule
\end{tabu}
\caption{Classification F1 scores of the models and the second rater annotations.}
\label{tab:f1_results}
\end{table*}

\begin{figure}[!thp]
\includegraphics[width=\columnwidth]{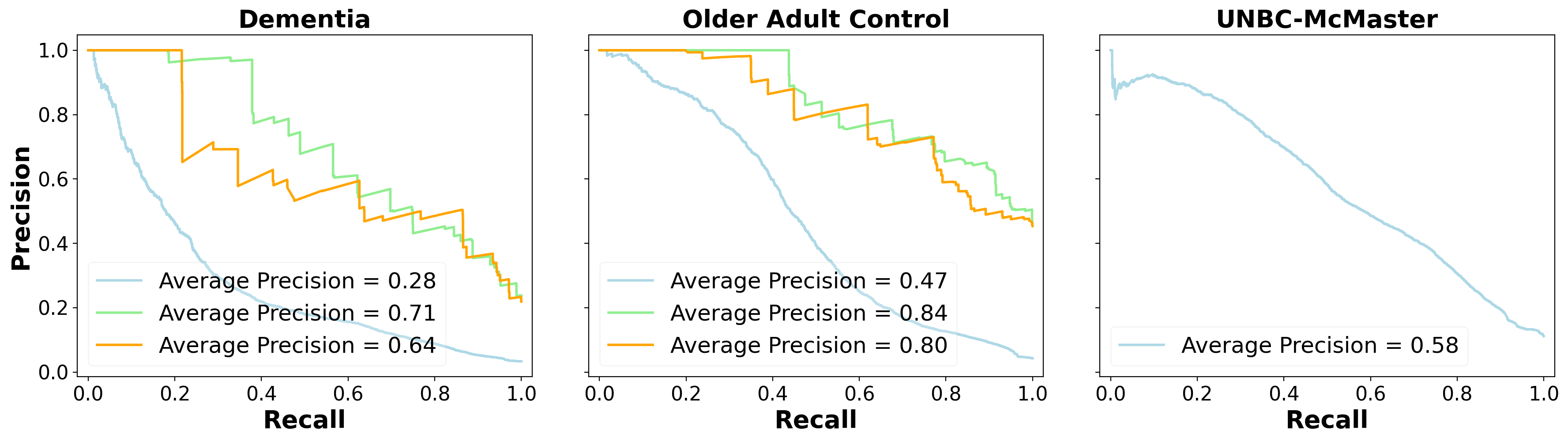}
\caption{Precision-Recall curves. Blue: per-frame prediction w/ contrastive training, Green: 20 second window prediction w/ contrastive training, Orange: 20 second window prediction w/o contrastive training.}
\label{fig:perc_rec_curves}
\end{figure}

\begin{figure}[!thp]
\includegraphics[width=\columnwidth]{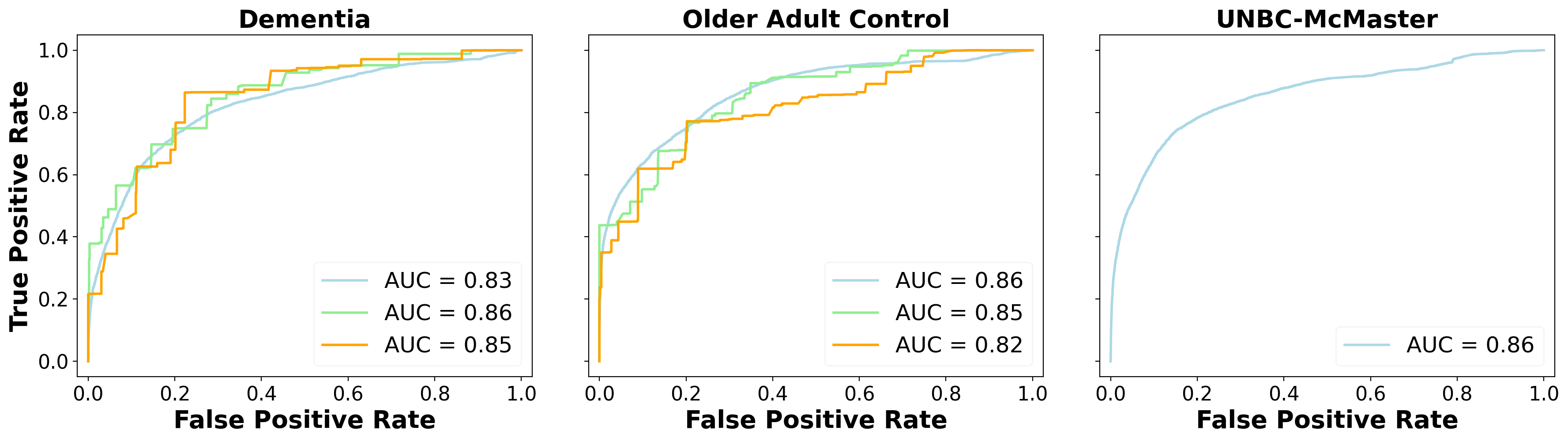}
\caption{ROC curves. Blue: per-frame prediction w/ contrastive training, Green: 20 second window prediction w/ contrastive training, Orange: 20 second window prediction w/o contrastive training.}
\label{fig:ROC}
\end{figure}

Figures \ref{fig:perc_rec_curves} and \ref{fig:ROC} depict classification precision-recall and ROC curves for per-frame and 20-second rolling window predictions. It is evident from these figures that rolling window aggregation results in a significant improvement in average precision. AUC for the dementia cohort also improves with aggregation. There are, however, no significant gains in AUC from window aggregation for the healthy control cohort. Also notable is the boost in average precision for the dementia group from contrastive training.\\

In order to assess the effect of multi-task learning, the model was trained to only estimate PSPI and not FACS and PACSLAC-II action units. Table~\ref{tab:multitask_ablation} reports the results. We notice performance degrades when multi-task training is eliminated. This effect is more prominent for cross-dataset performance.
Interestingly, without multi-task training, the model achieves better performance on the training set, which we attribute to overfitting. This suggests multitask training prevents the model from over-fitting the training data, and explains the improved cross-dataset performance.
\begin{table}[!thp]
\begin{tabularx}{\columnwidth}{l|>{\RaggedRight}X>{\RaggedRight}X>{\RaggedRight}X}
\toprule
 \diagbox{Dataset}{Output\\head}& Dementia & Older Adult Control & UNBC-McMaster\\
\midrule
Dementia &0.45 / 0.48 & 0.37 / 0.49 & 0.33 / 0.48 \\
Older adult control &0.51 / 0.54 & 0.61 / 0.58 & 0.52 / 0.57 \\
UNBC-McMaster&0.65 / 0.69& 0.68 / 0.70 &0.68 / 0.69\\
\bottomrule
\end{tabularx}
\caption{Pearson correlations for per-frame PSPI predictions; without/with multi-task training. Off-diagonal entries reflect cross-dataset performance.}
\label{tab:multitask_ablation}
\end{table}

\begin{figure}[!thp]
\centerline{\includegraphics[width=0.45\textwidth]{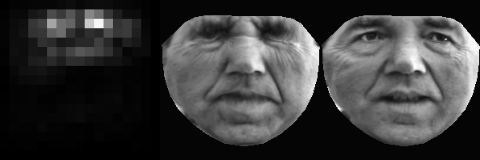}
\includegraphics[width=0.45\textwidth]{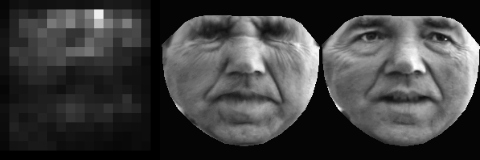}}
\caption{Top: saliency map generated from a model w/o contrastive training. Bottom: saliency map generated for the same sample from a model w/ contrastive training. Sample PSPI is 12.}
\label{fig:saliency_sample}
\end{figure}

\begin{figure}
  \begin{minipage}[c]{0.6\textwidth}
    \includegraphics[width=5cm]{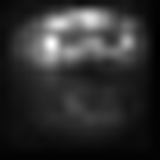}
    \includegraphics[width=5cm]{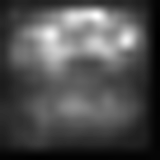}
  \end{minipage}\hfill
  \begin{minipage}[l]{0.4\textwidth}
    \caption{
        Mean of the saliency maps of the top 500 highest pain samples. Left: generated from a model w/o contrastive training. Right: generated from a model w/ contrastive training.
    } \label{fig:saliency_mean}
  \end{minipage}
\end{figure}

To get a qualitative feel for how contrastive training influences the network, we plot saliency maps following Selvaraju~et~al.~\cite{Selvaraju2019}. 
Figure~\ref{fig:saliency_sample} depicts the saliency maps of the same sample (from the UNBC dataset) with high pain score (PSPI\,=\,12). The high pain frame is shown in the middle column and a reference frame  (PSPI\,=\,0) is shown on the right column. Upper and lower left images show the saliency map generated from the model without and with contrastive training, respectively. We can see the area around the mouth is more salient for the model with contrastive training. This comports with expert field knowledge that movements in the mouth region are indicative of pain.
Figure \ref{fig:saliency_mean} is generated by averaging the saliency maps of the top 500 samples according to their PSPI scores. Again, we can see the mouth area is more salient to the model with contrastive training (right).

\subsubsection*{A Perspective on Contrastive Training}
The proposed model has 39 outputs in total. The outputs correspond to individual FACS and PACSLAC-II action units and PSPI for UofR control, UofR dementia, and UNBC.
The weights of the last linear layer (\(W_{fc}\)) map a 200-dimensional feature vector to these 39 outputs. These 200-dimensional weight vectors (i.e. columns of  \(W_{fc}\)) can be thought of as a basis set of (at most) rank 39 in a 200 dimensional space. The proposed contrastive loss jointly trains both \(f\) and \(W_{fc}\) (the basis vectors) such that feature vectors computed for OOD samples will lie outside of the column space of \(W_{fc}\), and feature vectors computed for ID samples lie in that column space.  
In contrast to Grathwohl~et~al.~\cite{Grathwohl2019}, who rely on the formulation of the softmax function to formalize their interpretation, here we show softmax is not necessary to do contrastive training. This disentanglement from a non-linear activation function, allows us to do contrastive training even for the for the middle layers of any neural network.\\


\section{Ethical Considerations}
\label{ethic}
\subsection{Equity and Algorithmic Bias}

Buolamwini and Gebru famously demonstrated that the performance of commercially available facial analysis software was significantly dependent on skin tone and gender~\cite{pmlr-v81-buolamwini18a}. Specifically, performance was as much as 34 percentage points worse on faces of darker skinned females than it was on faces of lighter skinned males. Our recent work has shown a similar bias in performance when facial analysis models are evaluated on faces of older adults with a physical or cognitive disability. We showed, for instance, that pre-trained facial landmark detection models and pre-trained facial action unit detection models performed significantly worse when evaluated on faces of older adults with dementia vs. cognitively healthy older adults~\cite{Asgarian_2019_CVPR_Workshops, 8643365}. A similar bias was observed when models were evaluated on faces of individuals with facial palsy~\cite{doi:10.1089/fpsam.2019.29000.gua,Guarin2020}, amyotrophic lateral sclerosis (ALS), and post-stroke~\cite{neuroface}. This is not surprising, as the dataset used to train these models consists primarily of younger adults and virtually entirely of healthy individuals. However, the existence of this bias places a major limitation on the application of existing computer vision models in healthcare settings. In the specific case of ambient pain monitoring in LTC homes, equitable access to quality care and pain management demands the same level of model performance regardless of race, gender, and underlying medical condition, especially the particular use case of this research, dementia. 

In this paper we presented the first fully automated computer vision pain detection model to be trained and evaluated on a large corpus of video data from older adults with dementia. In the analysis, we presented results separately for the dementia cohort and older adult controls so differences in performance could be studied. When evaluating per-frame regression results, the gap in performance between the two groups was large, but aggregating over a length of time closed the gap (Table~\ref{tab:pearson_results}). 

We do not have ethnicity/race breakdown for participants in either of the two datasets used, so sensitivity to race cannot be evaluated and remains the topic of future work. Participants' gender, however, is known and we can evaluate performance separately for male and female participants. The model (with contrastive training, evaluated per-frame, and using 5 reference frames from the same person) obtained a Pearson correlation of 0.48 in the dementia cohort of the UofR dataset (Table~\ref{tab:3_results}). Evaluating performance separately for male and female participants, the model obtained a Pearson correlation of 0.46 on male faces and 0.50 on female faces. While the difference is relatively small, future work should focus on reducing this performance gap. We note that the number of female face image frames in the UofR dataset was more than twice that of male faces; so collecting more data from male participants could be explored as a potential solution for reducing the gap. 

\subsection{Privacy} 

The use of ambient systems to monitor behaviour in LTC brings up immediate concerns about privacy. The overall vision is that the final automated system will be deployed onboard, e.g.\ using a Google Coral or an NVIDIA Jetson Nano, so image capture and processing is performed onboard and in real-time. This will remove the need for storage and will alleviate privacy concerns related to the viewing of videos by individuals who lack the necessary permissions to view private information. Previous studies have shown that monitoring technologies can be viewed as acceptable by older adults when the benefits outweigh privacy concerns~\cite{Mihailidis2008}.  In this case, potential loss of privacy could be outweighed by the benefits gained through prompt and effective management of pain among LTC residents.

\begin{table*}[!thb]
\centering
\resizebox{\textwidth}{!}{%
\begin{tiny}
\begin{tabular}{l|llll|llll|llll|llll}
\toprule
{ \textbf{Test}}  & \multicolumn{8}{l}{{ \textbf{VAS and PSPI}}}& \multicolumn{8}{l}{{ \textbf{Observer ratings and PSPI}}} \\ \hline
{ }  & \multicolumn{4}{l}{{ \textbf{Active tests}}}  & \multicolumn{4}{l}{{ \textbf{Passive tests}}} & \multicolumn{4}{l}{{ \textbf{Active tests}}}  & \multicolumn{4}{l}{{ \textbf{Passive tests}}} \\
{ }  & { \textbf{r}}  & { \textbf{AUC}}& { \textbf{mq}}  & { \textbf{Crit}} & { \textbf{r}}  & { \textbf{AUC}}& { \textbf{mq}}  & { \textbf{Crit}} & { \textbf{r}}  & { \textbf{AUC}}& { \textbf{mq}}  & { \textbf{Crit}} & { \textbf{r}}  & { \textbf{AUC}}& { \textbf{mq}}  & { \textbf{Crit}} \\
{ \textbf{Abduction1}}& { \textbf{.39 $\ast$}} & { .65}& { .31}& { 3}& { \textbf{.46 $\ast$}} & { .59}& { .46}& { 1}& { \textbf{.70 $\ast$}} & { \textbf{.83 $\ast$}} & { \textbf{.75}} & { 2}& { \textbf{.71 $\ast$}} & { \textbf{.86 $\ast$}} & { \textbf{.80}} & { 3}\\
{ \textbf{Flexion1}} & { \textbf{.36 $\ast$}} & { \textbf{.86 $\ast$}} & { \textbf{.75}} & { 4}& { \textbf{.20} $\ddagger$}   & { .69}& { .29}& { 5}& { \textbf{.67 $\ast$}} & { \textbf{.91 $\ast$}} & { \textbf{.84}} & { 2}& { \textbf{.70 $\ast$}} & { \textbf{.87 $\ast$}} & { \textbf{.79}} & { 3}\\
{ \textbf{Internal rotation1}} & { \textbf{.38 $\ast$}} & { .23}& { -.04}& { ?}& { \textbf{.37 $\ast$}} & { \textbf{.86 $\ast$}} & { \textbf{.80}} & { 5}& { \textbf{.67 $\ast$}} & { \textbf{.75 $\ast$}} & { \textbf{.57}} & { 3}& { \textbf{.70 $\ast$}} & { \textbf{.92 $\ast$}} & { \textbf{.84}} & { 2}\\
{ \textbf{External rotation1}} & { \textbf{.28 $\dagger$}}  & { .18}& { -.03}& { ?}& { \textbf{.32 $\ast$}} & { .51}& { .40}& { 1}& { \textbf{.53 $\ast$}} & { \textbf{.77 $\ast$}} & { \textbf{.63}} & { 2}& { \textbf{.76 $\ast$}} & { \textbf{.91 $\ast$}} & { \textbf{.85}} & { 4}\\
{ \textbf{Abduction2}}& { \textbf{.38 $\ast$}} & { .32}& { .12}& { ?}& { \textbf{.35 $\dagger$}}  & { \textbf{.71 $\ast$}} & { \textbf{.60}} & { 2}& { \textbf{.67 $\ast$}} & { \textbf{.83 $\ast$}} & { \textbf{.75}} & { 2}& { \textbf{.68 $\ast$}} & { \textbf{.73 $\dagger$}}  & { \textbf{.57}} & { 2}\\
{ \textbf{Flexion2}} & { \textbf{.37 $\ast$}} & { .68}& { .14}& { 4}& { \textbf{.32 $\dagger$}} & { .59} & { .23}& { 2} & { \textbf{.65 $\ast$}} & { \textbf{.88 $\ast$}} & { \textbf{.79}} & { 2}& { \textbf{.65 $\ast$}} & { \textbf{.77 $\dagger$}}  & { \textbf{.61}} & { 1}\\
{ \textbf{Internal rotation2}} & { \textbf{.28 $\dagger$}}  & { .61}& { .11}& { 7}& { .13}& { .46}  & { .38}& { 5} & { \textbf{.75 $\ast$}} & { \textbf{.88 $\ast$}} & { \textbf{.77}} & { 2}& { \textbf{.58 $\ast$}} & { \textbf{.80 $\ast$}} & { \textbf{.69}} & { 3}\\
{ \textbf{External rotation2}} & { \textbf{.27 $\dagger$}}  & { \textbf{.92 $\ast$}} & { \textbf{.87}} & { 5}& { \textbf{.33 $\ast$}} & {  .49}   & { .30}& { 4} & { \textbf{.58 $\ast$}} & { \textbf{.81 $\ast$}} & { \textbf{.62}} & { 2}& { \textbf{.68 $\ast$}} & { \textbf{.83 $\ast$}} & { \textbf{.70}} & { 2}\\
{ \textbf{~~~~~Average} }  & { \textbf{.34}}& { \textbf{.55}}& { \textbf{.28}} & { } & { \textbf{.31}}& { .67}& { .51}& { } & { \textbf{.65}}& { \textbf{.83}}& { \textbf{.72}} & { } & { \textbf{.68}}& { \textbf{.84}}& { \textbf{.73}} & { } \\ 
\bottomrule
\multicolumn{17}{l}{ $\ast$ p$<$.001,  $\dagger$ p$<$.01, $\ddagger$ p$<$.05, ? indicates a statistically insignificant model from which no criterion can be specified}\\
\multicolumn{17}{l}{\textbf{Boldface} identifies parameters meeting criteria for statistical significance.}\\
\multicolumn{17}{l}{VAS = visual analogue scale; AUC = area under the curve; mq = model quality; Crit = ROC suggested criterion.}\\
\multicolumn{17}{l}{ROC analysis for relationship between pain as defined by VAS criteria and observer ratings and PSPI scores. Optimal PSPI score for establishing cutoffs are identified in the Crit column.}
\end{tabular}%
\end{tiny}
}
\vspace{-0.25em}
\caption{Metrics for the self-report VAS ratings and PSPI scores on the left; and observer ratings and PSPI scores on the right.}
\label{tab:rationale}
\vspace{-1em}
\end{table*}

\section{Conclusion and Future Work}
\label{conclusion}
In this paper, we (1) developed the first fully automated computer vision model capable of estimating the level of pain based on facial expressions of older adults with dementia. We (2) proposed a new neural network architecture which compares two images from the same person and estimates the difference in the pain level in them. This is based on the observation that human annotators implicitly adjust their judgements to a person's resting facial expression. We also (3) introduced a contrastive training method where the model is trained to discriminate between in-distribution and out-of-distribution samples in addition to its default objective without changes to architecture. This contrastive training can be used with any neural network regardless of its architecture, and our results suggest that it may help with generalization. 

In the experimental evaluation with a large corpus of video data the resulting model outperformed existing baselines by a wide margin. Nevertheless, it is evident that there is still a large gap between human and machine performance on this task. This is specifically true for the UofR dataset, where the video recording setting was less controlled and more in line with what one can expect in an unobtrusive monitoring setting. 
The development of strong computer vision models in the past decade has been spearheaded by supervised deep learning which relies on the availability of a massive amount of labeled data. Collecting a large amount of clinical data, e.g.\ from a dementia population, is challenging, time consuming, and expensive due to difficulties in participant recruitment, obtaining informed consent from a vulnerable population with a cognitive disability, and the need for trained clinical staff to assist with data collection. For a task such as FACS-based pain detection, the barrier is even higher due to the great effort and resources required for manual FACS coding. 

Even though the currently available FACS coded datasets (i.e.\ UNBC-McMaster) come close to classic computer vision datasets (e.g.\ CIFAR10) in terms of the number of frames, they still lack the variability of object classification datasets. This is because the images are frames from videos, and videos have low frame-to-frame variability and are collected from a small number of participants, e.g.\ N=25 in the case of the publicly available portion of the UNBC-McMaster dataset. This makes the effective size of these datasets much smaller. This, added to the fact that the datasets are very imbalanced, means there are only a handful of frames with high pain for the models to learn from. It also makes large models prone to overfitting to the specific faces in the dataset.
All of this, we believe, makes automatic pain detection a suitable venue for leveraging unsupervised training methods. This is a line that we will pursue in our future research.

\appendices
\section{Rationale for selection of pain criterion}
\label{appx:threshold}

The task is to identify a PSPI score that is likely to be able to meaningfully identify pain in the behavioural stream. The UNBC-McMaster database has data that could help identify a criterion empirically.
In the complete UNBC-McMaster, approximately 130 people went through range of motion tests to activate both shoulders—one affected by pain, the other unaffected. There were four tests performed twice actively and twice passively, yielding 16 separate experiments in effect. During each test, participants rated the maximum pain using a 10-cm visual analogue scale (VAS), providing a self-report measure of how painful it was. An independent observer also viewed the video of each test, rating how painful it was on a 0-5 scale of pain intensity. Each test was also scored for pain expression intensity using PSPI.

An empirical criterion for identifying pain on the PSPI scale can be suggested by validating it against one or the other of the self-report or observer-rating measures, so long as there is reason to believe that the PSPI scale is related to them. Table~\ref{tab:rationale} presents the Pearson correlations between self-report VAS ratings and PSPI scores and also between the observer ratings and PSPI scores on the 16 pain experiments. In every case but one (passive internal rotation retest - VAS correlation), the PSPI scores and the validating measures are significantly correlated. Correlations against the observer ratings are strong and substantially higher than correlations against VAS scores. That raises the problem of what parameter to use to identify pain on those measures. Implicitly, a value of 0 on either measure should, in principle, mean no pain, but natural variability and error need to be taken into account. A conservative approach would be to take that variation into account and specify a value that one could assert with reasonable certainty would imply no pain. 

In the original study, the tests were conducted on the affected and the unaffected side. Challenging the affected side should produce pain at some level at least some of the time and this should be evident in both the self-reports and the observer ratings. Challenging the unaffected side should, in general, not be painful, but there will be variability associated with the test and the individual.
In line with this, a conservative approach to establishing what is no pain would be to take the average self-report or observer rating of the tests that should not be painful (unaffected side tests) and establish the 99\% confidence interval around it. Ratings outside that interval can be taken to mean that it is likely that sometime is truly in pain and this can establish a criterion by which the ability of the PSPI score to detect pain can be evaluated.

ROC analyses was run separately for each of the 16 experiments, specifying a VAS of 5 or higher or an observer rating score of 3 or higher to define pain. The results are in Table~\ref{tab:rationale}. In 11 of 16 cases (active abduction test, active flexion test and retest, active internal rotation retest, active external rotation retest, passive abduction test and retest, passive flexion test and retest, passive internal rotation test and passive external rotation test), the analysis based on VAS scores yielded AUC values greater than 0.5. Of those, four were statistically significantly different from 0.5 and achieved a model quality index greater than 0.5. Suggested PSPI scores across active and passive tests for defining pain (in the `Crit' column) varied between 1 and 7, with no clear consensual value. In the analyses based on the observer ratings, all AUC values substantially exceeded 0.5 and were significant and all the model quality indices were good. On 7 of 8 active tests, the recommended PSPI criterion for defining pain was 2. For passive tests, there was more variability in the recommended criterion. They average to 2.5.
Given the much higher correlations between PSPI scores and observer rating scores and the clearly superior metric properties when using observer ratings as a criterion, the recommendations based on those analyses are preferable. The results suggest a criterion of 2 or 3 for using PSPI scores in developing the algorithm.

\section{Results from additional experiments}
Table~\ref{tab:hist_eq_ablation} outlines results for a training run where histogram equalization was not used. Table~\ref{tab:3d_mesh} shows results from using 3D-mesh for frontalization instead of piecewise affine. In both experiments contrastive training was used.

Many other works (e.g.~\cite{5204259, Hammal2012, 7849133}) also report intraclass Correlation Coefficient~\cite{PMID:18839484} as a metric to judge the performance of their algorithm. Here we also report ICC(3,~1) in Table~\ref{tab:ICC}.

\begin{table}[!th]
\begin{tabularx}{\columnwidth}{l|>{\RaggedRight}X>{\RaggedRight}X>{\RaggedRight}X}
\toprule
 \diagbox{Dataset}{Output\\head}& Dementia & Older Adult Control & UNBC-McMaster\\
\midrule
Dementia &0.48 / 0.48 & 0.46 / 0.49 & 0.46 / 0.48 \\
Older adult control &0.51 / 0.54 & 0.56 / 0.58 & 0.55 / 0.57 \\
UNBC-McMaster&0.66 / 0.69& 0.70 / 0.70 &0.68 / 0.69\\
\bottomrule
\end{tabularx}
\caption{Pearson correlations for per-frame PSPI predictions; without/with histogram equalization. Off-diagonal entries reflect cross-dataset performance.}
\label{tab:hist_eq_ablation}
\end{table}

\begin{table}[!th]
\begin{tabularx}{\columnwidth}{l|>{\RaggedRight}X>{\RaggedRight}X>{\RaggedRight}X}
\toprule
 \diagbox{Dataset}{Output\\head}& Dementia & Older Adult Control & UNBC-McMaster\\
\midrule
Dementia &0.36 / 0.48 & 0.38 / 0.49 & 0.36 / 0.48 \\
Older adult control &0.53 / 0.54 & 0.54 / 0.58 & 0.53 / 0.57 \\
UNBC-McMaster&0.67 / 0.69& 0.67 / 0.70 &0.67 / 0.69\\
\bottomrule
\end{tabularx}
\caption{Pearson correlations for per-frame PSPI predictions; 3D-mesh/piecewise-affine frontalization method. Off-diagonal entries reflect cross-dataset performance.}
\label{tab:3d_mesh}
\end{table}

\begin{table}[!th]
\begin{tabularx}{\columnwidth}{l|>{\RaggedRight}X>{\RaggedRight}X>{\RaggedRight}X}
\toprule
 \diagbox{Dataset}{Output\\head}& Dementia & Older Adult Control & UNBC-McMaster\\
\midrule
Dementia &0.34 & 0.35 & 0.31 \\
Older adult control &0.41& 0.47& 0.40 \\
UNBC-McMaster&0.53& 0.56 &0.59\\
\bottomrule
\end{tabularx}
\caption{Intraclass Correlation Coefficient (specifically ICC(3, 1)) for per-frame PSPI predictions of the contrastively trained model . Off-diagonal entries reflect cross-dataset performance.}
\label{tab:ICC}
\end{table}


\bibliographystyle{unsrt}
\bibliography{export_2020-5-14, extra, export_2020-7-10}

\end{document}